\documentclass[11pt,a4paper]{article}
\usepackage[hyphens]{url}
\usepackage[hyperref]{include/emnlp-ijcnlp-2019}
\hypersetup{breaklinks=true}
\usepackage{times}
\usepackage{latexsym}

\aclfinalcopy %

\usepackage{amsmath}

\usepackage{cmap} 

\usepackage[utf8]{inputenc}
\usepackage[LGR,T1]{fontenc}
\usepackage[greek,english]{babel}

\usepackage{cleveref}

\usepackage{graphicx}

\usepackage{xspace}

\usepackage{etoolbox}
\AtBeginEnvironment{thebibliography}{\interlinepenalty=10000}

\newcommand{\modelname}{\textsc{Pythia}\xspace}
\newcommand{\datasetname}{\textsc{PHI\nobreakdash-ML}\xspace}
\newcommand{\minisec}[1]{\textbf{#1}}

\title{Restoring ancient text using deep learning:\\
a case study on Greek epigraphy}

\newcommand\authorspace{\hspace{1em}}
\author{
  Yannis Assael$^{1,2,*}$,\authorspace
    Thea Sommerschield$^{1,}$\thanks{\enspace These authors contributed equally to this work.}\enspace,\authorspace
    Jonathan Prag$^1$ \\
    University of Oxford$^1$, DeepMind$^2$
}

\begin{document}

\maketitle
\begin{abstract}
Ancient History relies on disciplines such as Epigraphy, the study of ancient inscribed texts, for evidence of the recorded past. However, these texts, ``inscriptions'', are often damaged over the centuries, and illegible parts of the text must be restored by specialists, known as epigraphists.
This work presents \modelname, the first ancient text restoration model that recovers missing characters from a damaged text input using deep neural networks.
Its architecture is carefully designed to handle long-term context information, and deal efficiently with missing or corrupted character and word representations. 
To train it, we wrote a non-trivial pipeline to convert PHI, the largest digital corpus of ancient Greek inscriptions, to machine actionable text, which we call \datasetname.
On \datasetname, \modelname's predictions achieve a $30.1\%$ character error rate, compared to the $57.3\%$ of human epigraphists. Moreover, in $73.5\%$ of cases the ground-truth sequence was among the Top-$20$ hypotheses of \modelname, which effectively demonstrates the impact of this assistive method on the field of digital epigraphy, and sets the state-of-the-art in ancient text restoration.

\end{abstract}

\section{Introduction}
\label{sec:introduction}

\begin{figure}[!b]
\centering
\includegraphics[width=\linewidth]{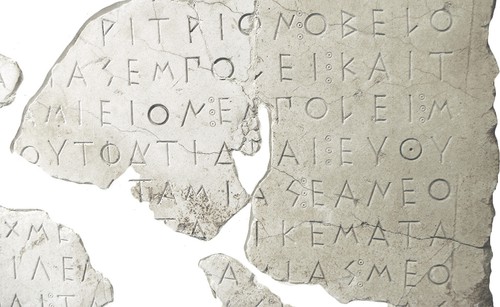}
\caption{%
Damaged inscription: a decree concerning the Acropolis of Athens ($485/4$ BCE). \textit{IG} I$^{3}$ $4$B. (CC BY-SA 3.0, WikiMedia)}
\label{fig:damaged_inscription}
\end{figure}

One of the key sources for Ancient History is the discipline of epigraphy, which delivers firsthand evidence for the thought, society and history of ancient civilisations. Epigraphy is the study of documents, ``inscriptions'', written on a durable surface (stone, ceramic, metal) by individuals, groups and institutions of the past \cite{davies2012epigraphy}.
Only a small minority of surviving inscriptions are fully legible and complete, as many have been damaged in time (\Cref{fig:damaged_inscription}). An epigraphist must then hypothesise how much text is missing, and what it might have originally been. These hypotheses are called ``restorations'' \cite{bodel2012epigraphy}. The present work offers a fully automated aid to the epigraphist's restoration task.

Restoring text is a complex and time-consuming task \cite{woodhead1967study,mattingly1996athenian}. Epigraphists rely on accessing vast repositories of information to find textual and contextual ``parallels'' (recurring expressions in similar documents). These repositories primarily consist in a researcher’s mnemonic repertoire of such parallels, and in digital corpora for performing ``string matching'' searches \cite{PHI,ECDS}. 
However, minor differences in the search query can exclude or obfuscate relevant results, making it hard to estimate the true probability distribution of possible restorations.
To the best of our knowledge, this is the first work to bypass the constraints of current epigraphic methods by means of a fully automated deep learning model, \modelname, which aids the task of ancient text restoration. It is supplemented by \datasetname, an epigraphic dataset of a machine actionable text.
\modelname takes as input a sequence of damaged text, and is trained to predict character sequences comprising the hypothesised restorations.
It works both at a character- and a word-level, thereby effectively handling incomplete or missing words.
\modelname can furthermore be used by all disciplines dealing with ancient texts (philology, papyrology, codicology) and applies to any language (ancient or modern).
To aid and encourage future research, \modelname and \datasetname have been open-sourced at \url{http://github.com/sommerschield/ancient-text-restoration}.

\section{Related work}
\label{sec:background}

Natural language processing (NLP) has dealt with tasks akin to text restoration. Indeed, standard count-based n-gram language models (LM) share with epigraphists the ``parallel-finding'' approach. 
N-gram models are outperformed by neural language models, which operate at a word-level \citep{mikolov2010recurrent,mikolov2011empirical}, at a subword- or character-level \citep{sutskever2011generating,mikolov2012subword,botha2014compositional}, or a combination of both, known as character-aware language models \citep{miyamoto2016gated,kim2016character,hwang2017character}.
Despite our efforts to include BERT \citep{devlin2018bert} in our evaluation, we found that the excessive resources required did not allow for training on a single GPU.
Text restoration also shares similarities with machine reading comprehension \citep{hermann2015teaching,kovcisky2018narrativeqa}, and cloze deletion tests \citep{hill2015goldilocks,bajgar2016embracing,fedus2018maskgan,xie2018large,zhang2018effective}. 
Although word-level language modelling is capable of capturing context information more efficiently than character-level alternatives, damaged inscriptions preserve only limited parts of words, complicating the learning of representations. To overcome this issue, \modelname works simultaneously at both a character- and a word-level, thereby capturing long-term dependencies (``context information'').

Finally, several works have used machine learning to study ancient inscriptions, focusing on assistive tools \citep{roued2010towards},
optical character recognition and visual analysis \citep{terras2006image,garz2014hisdoc,soumya2014classification,shaus2017computer,hussien2015optical,amato2016visual,can2016evaluating,suganya2017feature,palaniappan2017deep,avadesh2018optical}, writer identification \citep{tracy2009study,panagopoulos2009automatic,faigenbaum2016algorithmic}, text analysis \citep{rao2009markov,rao2009entropic,rao2010entropy,yadav2010statistical,lee2010porting,vatri2018diorisis}, and machine translation \cite{page2017machine,luo2019neural}.

\section{Generating \textsc{\datasetname}}
\label{sec:dataset}
\label{subsec:phi}

Due to availability of digitised epigraphic corpora, \modelname has been trained on ancient Greek (henceforth, "AG") inscriptions, written in the ancient Greek language between $7$\textsuperscript{th} century BCE and $5$\textsuperscript{th} century CE.
We chose AG epigraphy as a case study for two reasons:
a) the variability of contents and context of the AG epigraphic record makes it an excellent challenge for NLP;
b) several digital AG textual corpora have been recently created, the largest ones being PHI \cite{PHI, Gawlinski} for epigraphy; Perseus \citep{Perseus} and First1KGreek \citep{opengreekandlatin} for ancient literary texts.

When restoring damaged AG inscriptions, the epigraphists' conjectures on the total number of missing characters are guided by grammatical and syntactical considerations, as well as by the reconstructed graphical layout of the inscription.
Conjectured missing characters are conventionally marked with hyphens, one hyphen equating to one missing character. Additionally, epigraphists traditionally convert edited texts to lower case and add punctuation and diacritics, which are generally absent from the original inscription. These conventions were also used in PHI.

Because human annotations in PHI were noisy and often syntactically inconsistent \cite{Inversen}, we wrote a pipeline to convert it into a machine actionable text.
We first computed the character frequencies and standardised the AG alphabet to include all core characters, including all accentuation ($147$ characters), numbers, spaces and punctuation marks. Two additional characters were introduced: `-' representing a missing character, and `?' signifying a character to be predicted.
Then we wrote regular expressions to replace all AG numerical notations appearing in the texts with $0$ to avoid numerical correlations, strip the remaining punctuation marks, remove the conventional epigraphical symbols surrounding certain characters (``Leiden Conventions''), and discard notes whose content was not in Greek. 
We then proceeded to clear human comments, fix the spacing and cases of duplicate punctuation, and filtered the resulting text so as to retain only the restricted alphabetical characters. The texts with fewer than $100$ characters were also discarded.
Lastly, we matched the number of missing characters with those conjectured by epigraphists, thereby converting the length value to an equal number of `-' symbols.

The resulting dataset is named \datasetname, and consists of more than $3.2$ million words (\Cref{tbl:dataset}). The inscriptions whose PHI IDs ended in \{3, 4\} (every inscription in PHI was assigned a unique identifier when the original corpus was created) were held out and used respectively as test and validation sets.

\begin{table}[htb]
\begin{center}
\begin{tabular}{|r|r|r|r|}
\hline \bf Split & \bf Inscriptions & \bf Words & \bf Chars \\ \hline
Train & $34,952$ & $2,792$k & $16,300$k \\
Valid & $2,826$ & $211$k & $1,230$k \\
Test & $2,949$ & $223$k & $1,298$k \\
\hline
\end{tabular}
\end{center}
\vspace{-1em}
\caption{Statistics for the \textsc{\datasetname} corpus.}
\label{tbl:dataset}
\vspace{-1em}
\end{table}
\section{Restoring text using \modelname}
\label{sec:model}

\begin{figure*}[tb]
\centering
\includegraphics[width=0.9\linewidth]{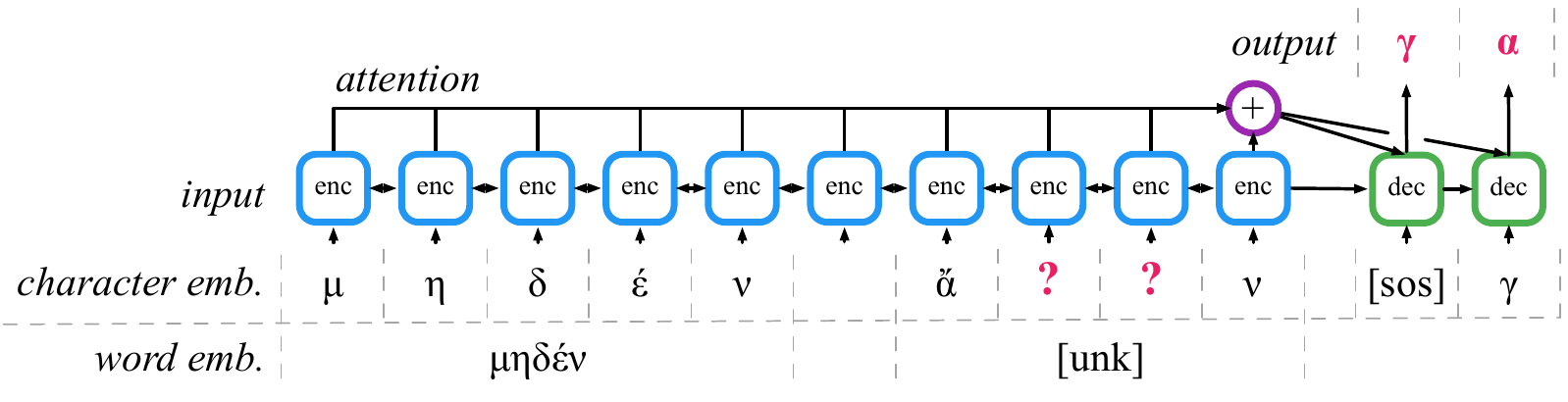}
\vspace{-0.5em}
\caption{\textsc{\modelname-Bi-Word} processing the phrase \textgreek{μηδέν ἄγαν} (mēdén ágan) ``nothing in excess'', a fabled maxim inscribed on Apollo's temple in Delphi. The letters ``\textgreek{γα}'' are the characters to be predicted, and are annotated with `?'. Since \textgreek{ἄ}??\textgreek{ν} is not a complete word, its embedding is treated as unknown (`unk'). The decoder outputs correctly~``\textgreek{γα}''.}
\label{fig:hermesarch}
\vspace{-0.5em}
\end{figure*}

\minisec{\modelname's architecture}
is a sequence-to-sequence \citep{sutskever2014sequence} based neural network architecture, consisting of a Long-Short Term Memory (LSTM) \citep{hochreiter1997long} encoder, an LSTM decoder, and an attention mechanism introduced by \citet{luong2015effective,bahdanau2014neural}.
The encoder takes an inscription text $x$ as input, where the symbol `-' denotes the missing characters, and `?' the blanks to be predicted. The input characters are first passed through a lookup table with learnable embedding vectors.
Next, the encoded sequence is used as input for the decoder, which is trained to predict the content of the `?' characters, as shown in \Cref{fig:hermesarch}.
Attention allows the decoder to ``attend'' to parts of the input sequence relevant to the current output, thus improving the modelling of long-term dependencies.
To further improve performance, we designed \modelname's encoder to take an additional input stream of word embeddings, as it is difficult to model the word-level context using only character-level information. Thus, we generated a list of the $100$k most frequent words appearing in \datasetname, and using a separate lookup table we concatenated at each time-step the embedding of each character with the embedding of the word it belongs to. Words that do not appear in the list, or that contain missing characters were mapped to `unk', an embedding for unknown words. \Cref{fig:hermesarch} illustrates \modelname processing the phrase \textgreek{μηδέν ἄγαν}.
Finally, to allow better modelling we used a bidirectional LSTM encoder and refer to this architecture as \textsc{\modelname-Bi-Word}. Further details are given in \Cref{sec:model-params}.

\minisec{Obtaining suggestions.} 
To better aid the epigraphist's task, \modelname returns multiple predictions as well as the level of confidence for each result, rather than a single prediction per text restoration. Specifically, we provide a set of the Top $20$ predictions decoded using beam search. %

\section{Experimental evaluation}
\label{sec:evaluation}

The ground-truths for incomplete epigraphic texts were lost over millennia. Consequently, in order to generate a ground-truth sequence, we artificially removed part of the input text and treated this as the ground-truth sequence.
On each training step we selected an inscription and sampled a start index and a length value $\in[100, 1000]$, and extracted the context text $x$, which was then used as input.
Within $x$, we sampled a new start index and length $\in[1, 10]$ to select the target sequence $y$; its characters' positions were replaced with the special symbol `?', which denotes the positions to be predicted.
The test and validation sets used the maximum context length. %
Beam search with a beam width of 100 was used to decode hypothesis.
To simplify comparisons, all AG accentuation was discarded, as inputting accents was time-consuming for the human evaluations described in the following paragraph. This decision did not noticeably influence the reported scores.

\subsection{Methods evaluated}
\textbf{\textsc{Ancient historian.}}
Because text restoration is an extremely time-consuming task even for an expert epigraphist, we set out to evaluate the difficulty of the restoration task at hand - and thereby judge the impact of our work - with the help of two doctoral students with epigraphical expertise.
The scholars were allowed to use the training set to search for ``parallels'', and made an average of $50$ restorations in $2$ hours, with a $57.3\%$ character error rate (CER).

\noindent\textbf{\textsc{LM philology.}} To evaluate the performance of a model using ``parallels'', we trained a LM.
Since large parts of the text are garbled, making complete words unidentifiable, and because BERT was not an option, the LM works at a character-level and is based on the setup of \citet{zaremba2015recurrent} (\Cref{sec:lm-params}). The LM was trained on two larger digital corpora of literary AG texts (``philology''), First1KGreek and Perseus, and evaluated on \textsc{\datasetname}.

\noindent\textbf{\textsc{LM philology \& epigraphy.}} LM jointly trained on First1KGreek, Perseus and \textsc{\datasetname}.

\noindent\textbf{\textsc{LM epigraphy.}} LM trained on \textsc{\datasetname}.

\noindent\textbf{\textsc{\modelname-Uni}.} An ablation architecture, using only characters as input and unidirectional LSTM.

\noindent\textbf{\textsc{\modelname-Bi}.} Similar to the \textsc{\modelname-Uni} ablation, but with a bidirectional LSTM.

\noindent\textbf{\textsc{\modelname-Bi-Word}.} This is our proposed model of choice, which uses a bidirectional LSTM and both characters and words as inputs.

\subsection{Results}

The aforementioned methods were evaluated using: a) the character error rate (CER) of the top prediction and the target sequence, b) the Top-20 accuracy score, where we ascertain whether the ground-truth sequence exists within the first 20 predictions.
The latter evaluates the effectiveness of \modelname as an assistive tool providing restoration suggestions to epigraphists.
As shown in \Cref{tbl:eval},
the ancient historians' restorations had a CER of $57.3\%$, which is telling of the difficulty of the task. 
The language model trained on epigraphic datasets performed slightly better, with a CER of $57.3\%$. Interestingly, the two attempts to use larger philological datasets performed worse. This is very likely due to a divergence in epigraphical and literary cultures.
The CER of the unidirectional \textsc{\modelname-Uni} and the bidirectional \textsc{\modelname-Bi} alternatives were $42.2\%$ and $32.5\%$ respectively. The top score was therefore achieved by the bidirectional \textsc{\modelname-Bi-Word}, which took both word and character embeddings as inputs, with a CER of $30.1\%$.
Furthermore, the ground-truth appeared among the $20$ most probable predictions of \textsc{\modelname-Bi-Word} $73.5\%$ of the times, which indicates that it could be a uniquely effective assistive tool.

\begin{table}[tb]
\begin{center}
\begin{tabular}{|l|r|r|r|}
\hline \bf Method & \bf CER & \bf Top-20 \\ \hline
Ancient Historian & $57.3\%$ & $-$ \\
\hline
LM Philology & $68.1\%$ & $26.0\%$ \\
LM Philology \& Epigraphy & $65.0\%$ & $28.8\%$ \\
LM Epigraphy & $52.7\%$ & $47.0\%$ \\
\hline
\textsc{\modelname-Uni} & $42.2\%$ & $60.6\%$ \\
\textsc{\modelname-Bi} & $32.5\%$ & $71.1\%$ \\
\textbf{\textsc{\modelname-Bi-Word}} & $\mathbf{30.1\%}$ & $\mathbf{73.5\%}$ \\
\hline
\end{tabular}
\end{center}
\vspace{-1em}
\caption{Predictive performance on \datasetname.}
\vspace{-1em}
\label{tbl:eval}
\end{table}

\subsection{The importance of context}
\label{sec:importance-context}

\begin{figure}[!b]
\centering
\includegraphics[width=\linewidth]{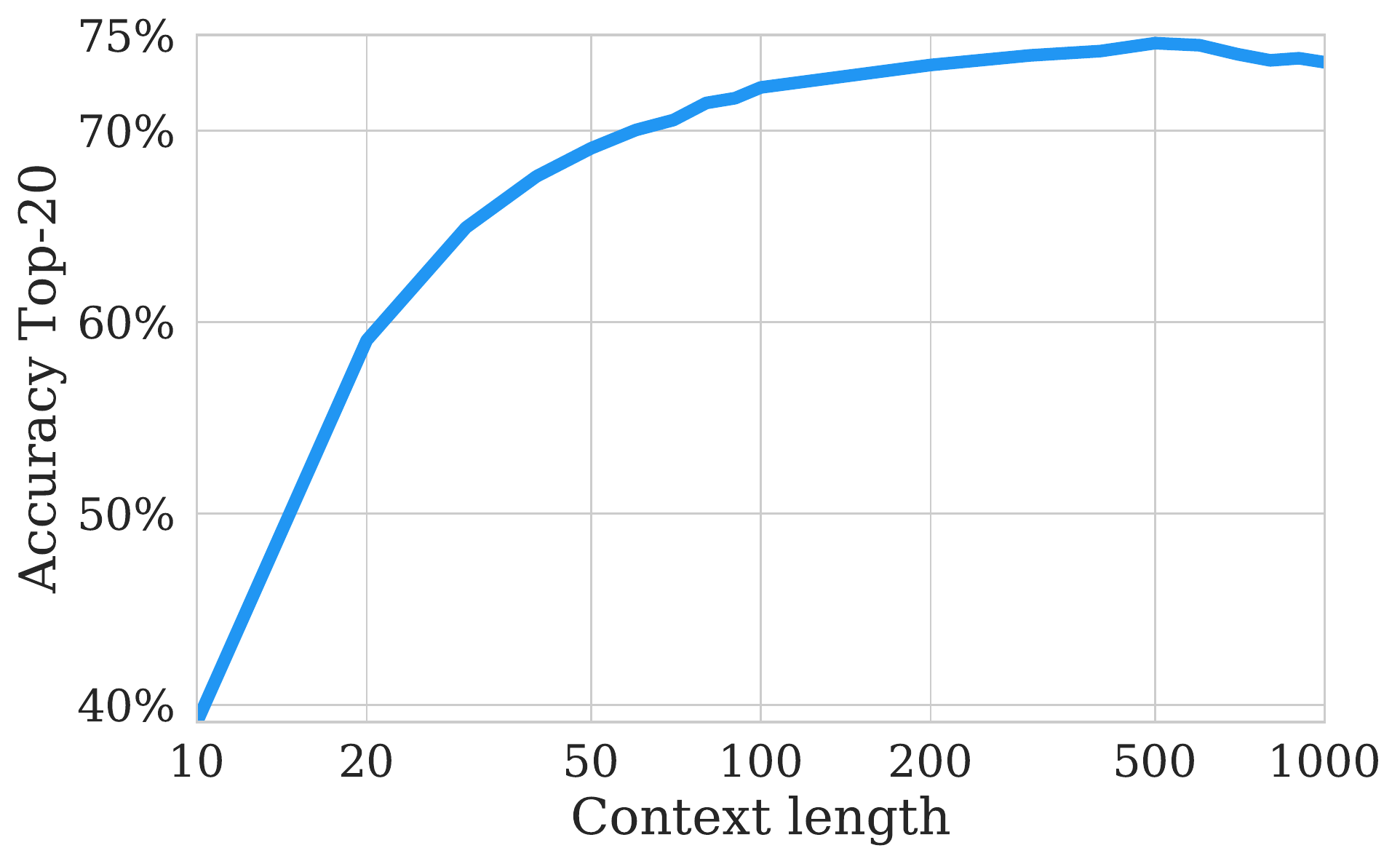}
\caption{Predictive performance of the \textsc{\modelname-Bi-Word} model under different context lengths.}
\label{fig:evalcontext}
\end{figure}

The presence of context information is a determining factor in the accuracy of epigraphic restorations.
We therefore evaluated the impact of different textual lengths acting as augmented context on the Top-20 accuracy measure of \textsc{\modelname}. As can be seen in \Cref{fig:evalcontext}, the correlation between the ``context length'' and the predictive performance of our model is positive. Specifically, the performance peaks around $500$ characters of context length. Furthermore, \Cref{fig:evalcontext} exemplifies the increased difficulty faced by the model when only a short context length (e.g. $20$ characters) is offered. The latter scenario recalls the similar difficulties encountered by string-matching and ``parallel'' search approaches, where the search queries would often be short.

\subsection{Visualising \modelname's attention}

\begin{figure}[b!]
\centering
\includegraphics[width=\linewidth]{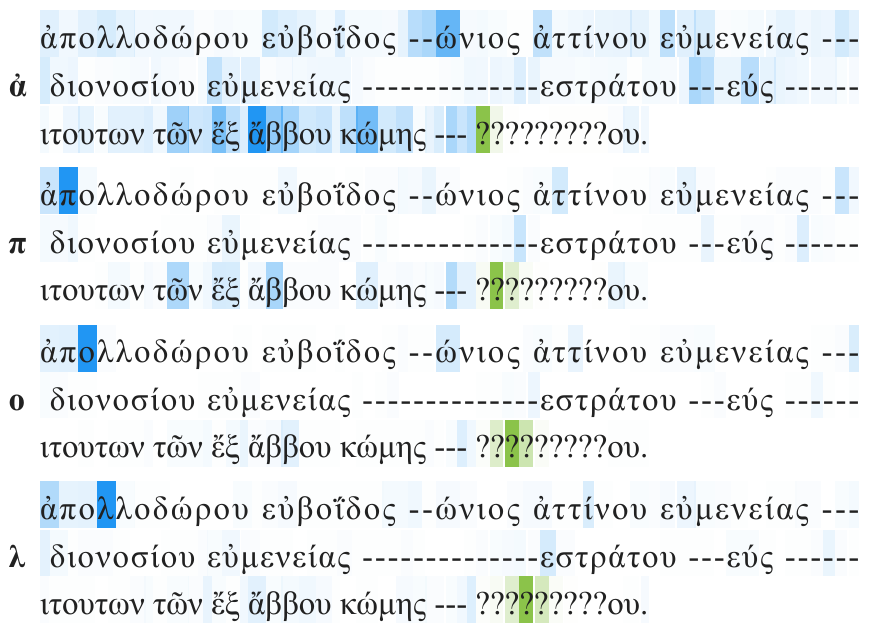}
\vspace{-1em}
\caption{Visualisation of the attention weights for the decoding of the first $4$ missing characters. The ground-truth text $y=$ \textgreek{ἀπολλοδώρ} appears in the input text, and \modelname attends to the relevant parts of the sequence.}
\label{fig:attnapol}
\end{figure}

We set up an example modifying lines b.$8$ - c.$5$ of the inscription MDAI(A) $32$ ($1907$) $428,275$ (PHI ID PH$316753$), to evaluate \modelname's receptiveness to context information and visualise the attention weights at each decoding step.
In the text of \Cref{fig:attnapol}, the last word is a Greek personal name ending in \textgreek{-ου}. We set \textgreek{ἀπολλοδώρου} (``Apollodorou'') as the personal name, and hid its first $9$ characters. This name was specifically chosen because it already appears within the input text. \Cref{fig:attnapol} illustrates the attention weights for decoding the first $4$ missing characters. To aid visualisation, the weights were separately scaled between $0$ and $1$ within the area of the characters to be predicted (`?') in green, and of the rest of the text in blue; the magnitude is represented by the colour intensity. As can be seen, \modelname is attending to the contextually-relevant parts of the text: specifically, \textgreek{ἀπολλοδώρου}. The name is correctly predicted.
As a litmus test, we substituted \textgreek{ἀπολλοδώρου} in the input text with another personal name of the same length: \textgreek{ἀρτεμιδώρου} (``Artemidorou''). The predicted sequence alters accordingly to \textgreek{ἀρτεμιδώρ}, thereby illustrating the importance of context in the prediction process.

\subsection{Restoring full texts}

\begin{figure}[bth!]
\centering
\includegraphics[width=\linewidth]{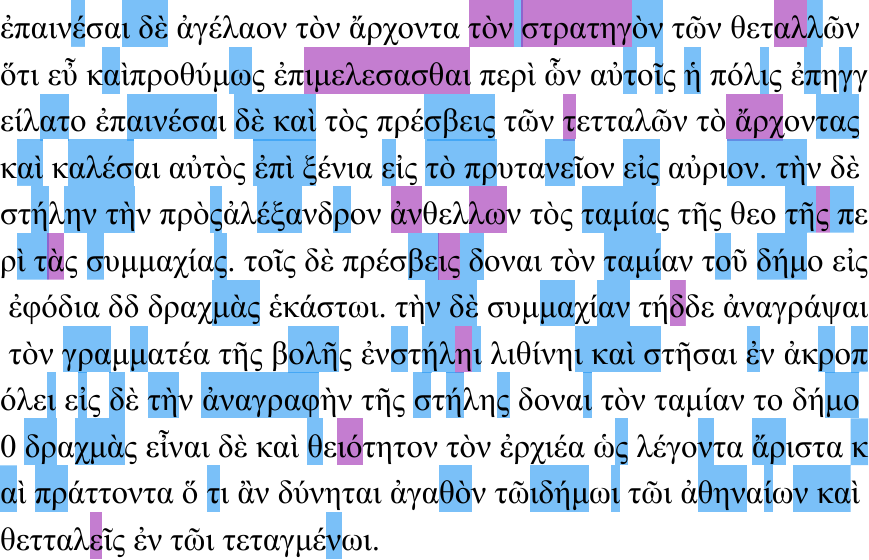}
\vspace{-1em}
\caption{Sample restoration of the inscription \textit{IG} II$^{2}$ $116$, lines $34$ - $48$. Restorations are in colour blue when correct, purple when incorrect.}
\label{fig:2234sm}
\vspace{-1em}
\end{figure}

We then applied \modelname iteratively in order to predict all the missing text of an AG inscription, comparing \modelname's predictions with an edition of reference \cite{rhodes2003greek}.
In \Cref{fig:2234sm} the correct restorations are highlighted in colour blue and erroneous ones are in purple. 
In a real-world scenario, \modelname would provide more than one hypothesis to the epigraphist. The ground-truth sequence did in fact exist within the Top-20 hypotheses in nearly all cases, illustrating the efficacy of such technologies when paired with human decision-making.

\section{Conclusions}
\label{sec:conclusions}

\modelname is the first ancient text restoration model of its kind.
Our experimental evaluation and ablation studies illustrate the validity of our design decisions, and illuminate the ways \modelname can assist, guide and advance the ancient historian's task - and digital humanities proper.
The combination of machine learning and epigraphy has the potential to impact meaningfully the study of inscribed textual cultures, both ancient and modern.
By open-sourcing \modelname, and \datasetname's processing pipeline, we hope to aid future research and inspire further interdisciplinary work.

\subsection*{Acknowledgements}
We would like to thank Leah Lazar for her valuable contributions; Brendan Shillingford, Misha Denil, \c{C}a\u{g}lar G\"{u}l\c{c}ehre, and Nando de Freitas for the helpful comments and discussions; and the Packard Humanities Institute for having made this data digitally available.

\bibliography{refs} 
\bibliographystyle{include/acl_natbib_nourl}

\clearpage
\onecolumn
\appendix

\section{\modelname training parameters}
\label{sec:model-params}
Both encoder and decoder of \modelname consist of 2-layers with $512$ hidden units. During training, we use dropout with probability $0.2$ and scheduled sampling with probability $0.5$ \citep{bengio2015scheduled}. 
All models were trained on an $8$-core machine with an NVIDIA $1080$ Ti graphics processing unit (GPU). The batch size was $32$ and the network weights were optimised using Adam \cite{kingma2014adam} with a learning rate of $10^{-3}$ and gradient clipping of $5$.

\section{LM training parameters}
\label{sec:lm-params}
The language modelling LSTM network consists of $2$-layers with $1024$ hidden units and an equally sized character embedding space. The parameters were trained using Adam with a learning rate of $2\cdot10^{-3}$, a decay of $0.95$, gradient norm clipping of $5$, and dropout probability $0.2$ for the inputs and the hidden layers.

\end{document}